\documentclass{article}
\usepackage[utf8]{inputenc}
\usepackage[nonatbib,final]{nips_2018}

\usepackage[T1]{fontenc}    
\usepackage{hyperref}       
\usepackage{url}            
\usepackage{amsfonts}       
\usepackage{nicefrac}       
\usepackage{microtype}      
\usepackage{booktabs}
\usepackage{array}

\usepackage[colorinlistoftodos]{todonotes}

\title{Predicting Language Recovery after
Stroke with Convolutional Networks on Stitched MRI}
\author{
\small 
Yusuf H. Roohani\\
  GlaxoSmithKline, Cambridge, MA \\
  \texttt{yusuf.x.roohani@gsk.com} \\
  \And
  Noor Sajid \\
  University College London \\
  \texttt{noor.sajid.18@ucl.ac.uk} \\
  \And
 Pranava Madhyastha \\
  Imperial College London\\
  \texttt{pranava@imperial.ac.uk} \\
  \And
  Cathy J. Price\\
  University College London \\
  \texttt{c.j.price@ucl.ac.uk} \\
  \And
  Thomas M. H. Hope \\
  University College London \\
  \texttt{t.hope@ucl.ac.uk} \\
}

\begin{document}
\maketitle
\begin{abstract}
One third of stroke survivors have language difficulties. Emerging evidence suggests that their likelihood of recovery depends mainly on the damage to language centers. Thus previous research for predicting language recovery post-stroke has focused on identifying damaged regions of the brain. In this paper, we introduce a novel method where we only make use of~\emph{stitched $2-$dimensional cross-sections} of raw MRI scans in a deep convolutional neural network setup to predict language recovery post-stroke. Our results show: a) the proposed model that only uses MRI scans has comparable performance to models that are dependent on lesion specific information; b) the features learned by our model are complementary to the lesion specific information and the combination of both appear to outperform previously reported results in similar settings. We further analyse the CNN model for understanding regions in brain that are responsible for arriving at these predictions using gradient based saliency maps. Our findings are in line with previous lesion studies.
\end{abstract}

\section{Introduction}

Stroke is one of the most common causes of disability. One third of stroke survivors leave the hospital with difficulties relating to cognitive and language understanding~\cite{seghier2016ploras}. This is known as aphasia or dysphasia in less severe cases. The likelihood of a patient to recover their language capabilities after stroke is thought to depend mainly on the proportion of damage to the brain and intensity of the initial symptoms ~\cite{plowman2012aphasia,cerebral2012}.  

Previous research has focused on the explicit use of brain structures derived from anatomically defined regions of the brain~\cite{hope2013ploras,hope2018ploras,price2010ploras,plowman2012aphasia,cross2007}. This, more often than not, will require specialists' knowledge of the brain. The usual features for prediction tend to make use of the proportion of damaged regions of the brain~\cite{price2010ploras,hope2013ploras}, commonly referred to as lesions, alongside demographic and behavioural features. However, outside the realm of predicting recovery post-stroke, Wilson el al.~2009~\cite{wilson2009nonploras} have proposed the use of principle component analysis for a more direct feature extraction process from MRI scans to predict primary progressive aphasia variants. 

The paper introduces a novel method that makes use of~\emph{stitched $2-$dimensional cross-sections} of raw MRI scans in a deep convolutional neural network setup. The results indicate that our proposal is able to predict language outcome post stroke with comparable performance to models that are dependent on expert derived lesion specific information. We measure language ability using Comprehensive Aphasia Test's (CAT) spoken picture description score ~\cite{aphasia2004}. This score is highly correlated with the prediction of language recovery ~\cite{hope2013ploras} and assesses the patient's ability to verbally describe a picture in three words or more. 

\begin{table*}[!htb]
\small
\begin{minipage}{.5\linewidth}
\centering
\begin{tabular}{@{}lcc@{}}\toprule
Features & Hope et al.~2013~&~Ours\\
\midrule
Baseline &  0.0* & 0.0\\
Img. Rep. & - & 0.53 \\
Demographic & - & 0.13\\
$+$Lesion* & 0.50* & 0.56\\
$+$Img. Rep. & - & \textbf{0.60}\\\bottomrule
\end{tabular}
\caption{R-squared}\label{tab:rsquare}
\end{minipage}\hfill
\begin{minipage}{.5\linewidth}
\centering
\begin{tabular}{@{}lcc@{}}\toprule
Features & Hope et al.~2018~&~Ours\\
\midrule
Baseline & - & 0.25\\
Img. Rep. & - & 0.74 \\
Demographic & - & 0.42\\
$+$ Lesion* & 0.73 & 0.75 \\
$+$ Img. Rep. & - & \textbf{0.78} \\\bottomrule
\end{tabular}
\caption{Pearson R's scores}\label{tab:correlation}
\end{minipage}
\caption{Comparison to previous  work~\cite{hope2013ploras,hope2018ploras}. *Note: previous approaches use different splits}
\end{table*}

\section{Related Work}
Recent work in the domain of predicting language recovery has focused on the relevance of imaging-based methods for better understanding language recovery post-stroke patients. Price et al.~2010 \cite{price2010ploras} introduce a data centric system 
that relies on structural MRI data in combination with behavioral data from standardised assessments, and demographic information to better predict individual outcomes and recovery post-stroke. Saur et al.~2010 \cite{saur2010brain} demonstrate the usefulness of language functional MRI activations to predict individual language outcomes obtained six months post stroke as a binary classification problem using support vector machines. Their work highlighted the importance of using imaging based methods since limiting the feature space only to age and current language deficit reduced the accuracy from $0.86$ to $0.62$. 

Hope et al.~2013~\cite{hope2013ploras} used PLORAS data to predict severity of language impairment, at the individual level, months or years after stroke onset. Their work was reliant on lesion identification techniques~\cite{seghier2008ploras} for converting the MRIs to \emph{anatomically defined regions of interest} that are destroyed~\cite{Eickhoff2005anatomy}. Their work also emphasized the importance of using more representative information about lesion location derived from MRIs. Their R-squared results increased from $0$ to $0.50$ with inclusion of finer-grained lesion-location data. Hope et al.~2018~\cite{hope2018ploras} again showcased the necessity of using MRI derived lesion information to predict language outcomes post stroke. However, in all these studies, they tend to rely on more statistical and expert driven methodologies on lesion extraction and language outcome prediction. In contrast, our proposed method relies directly on MRI scans. Zikic et al. ~2014 ~\cite{zikic2014miccai} have shown the success of using convolutional neural networks for brain tumor classification using a combination of 2d and 3d MRI scans as inputs. 



\section{PLORAS with CNN using Image Stitching}

We faced a few challenges with training a network directly on the $3D$ fMRI scans. Given the scarcity of data, we felt it would be difficult to effectively train $3D$ convolutional operations due to the higher number of free parameters required as compared to $2D$ ~\cite{classmri2018}. We instead choose to use axial cross sections to capture the maximum variation while minimizing the number of trainable parameters. 

However, analyzing $2$-dimensional slices individually was also not an option as the model would not be able to access vital contextual information across different scans. To overcome this, we proposed to \emph{stitch the $2D$ slices together} for each scan to create a single large $2D$ image for each scan. (Figure~\ref{fig:stitched} ). In this setup, the MRI scans followed a standard numbering system such that each voxel corresponded to the same location of the brain across different scans. Thus, we ensured that the individual image layers within each scan are in the same order and each pixel in the $2D$ stitched image physiologically matched every other image in the training set. We found that this setup was very helpful as it prevented the neural network from training on meaningless variation in the dataset.


\section{Experiments}
\paragraph{Data}
Our dataset comprises of stroke patients from the Predicting Language Outcome Recovery After Stroke (PLORAS) database~\cite{seghier2016ploras}. For each patient, we have demographic information, high resolution T1-weighted post-stroke MRI brain scans and associated CAT behavioral test results~\cite{aphasia2004}. PLORAS dataset contains a total of $1,858$ records from $835$ patients. There are $1,211$ unique assessments for spoken picture description scores and their associated MRI scan, of which $825$ entries were initial assessments, and the remaining $386$ were follow-ups. 

The training set contained $1,088$ patients, $348$ females and $740$ males, with an average age at stroke of $55.5$ (Q1-Q3: $46 - 64$). The spoken picture description outcomes were skewed distributed, ranging from $39$ to $75$. These scores were assessed from any time post-stroke, ranging from one day to $42$ years. The test set had $123$ patients with a balanced split between the classes. 

The following demographic features are included in our model a) years between stroke and scan, b) whether vision is affected, c) whether hearing is affected, d) gender, e) number of lesions, f) localisation of lesion; left and right, g) years of education, h) age at stroke  i) age since stroke and j) handedness. Alongside, we make use of expert derived lesion information. The baseline model (see Table~\ref{tab:rsquare}) includes only a), d), h) and j) as defined in Hope et al.~2013~\cite{hope2013ploras}.

\begin{figure}
\centering
\begin{minipage}{.5\textwidth}
  \centering
  \includegraphics[width=\linewidth,keepaspectratio]{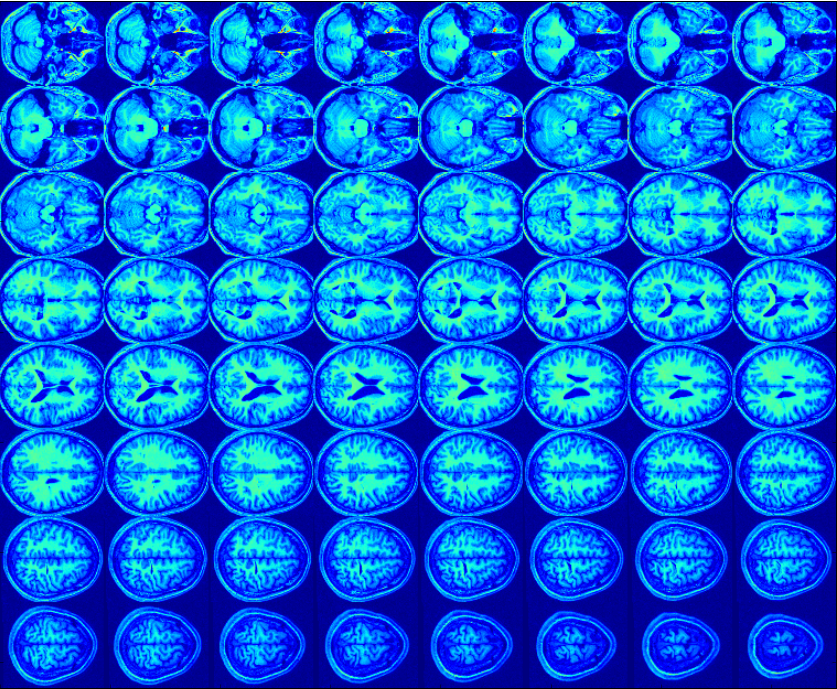}
\end{minipage}\hfill
\begin{minipage}{.5\textwidth}
  \centering
  \includegraphics[width=\linewidth,keepaspectratio]{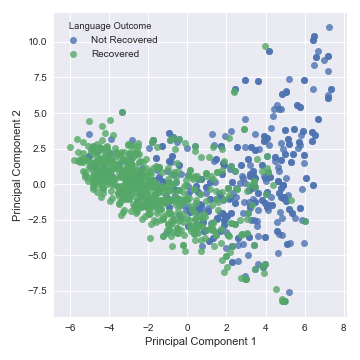}
\end{minipage}
\caption{Left: The 2-dimensional stitched MRI scans. Right: Visualization of the penultimate representation using PCA}
\label{fig:stitched}
\end{figure}

\paragraph{Setup}
We use $2D$ stitched images to train a convolutional neural network that classifies images as above or below the threshold (in our experiments this was $60$ based on Hope et al.~2018~\cite{hope2018ploras}).
Our model was composed of $869$ trainable parameters. The input to the model was a resized image of size $256{\times}256$. We trained the model using $5-$fold cross validation and the model achieved an average accuracy of $82\%$. We extracted the $64-$dimensional output from the final convolutional layer as image feature representation. 
In order to visualize our learned image feature representation, we project the data into the first two principle components. Figure~\ref{fig:stitched}(right) illustrates the ability of these features to distinguish between the classes within the validation set. In comparison with the demographic features, we see that our model has learned highly discriminative features. 

The extracted $64$-dimensional feature vector for each image was used to regress against the spoken picture description score along with the rest of the demographic features. 
For this purpose, we use a feed-forward neural network with $1$-hidden layer, adaptive moment estimation for optimisation and mean squared error as the loss. We also trained the MRI scans directly on a convolutional neural network regressor and found that the architecture has to be much more extensive to achieve comparable results and it does not allow for the inclusion of demographic features. 

\paragraph{Results and Observations} Our CNN classifier obtained an accuracy of $79\%$ on the held-out test set. We then use the features to train the feed-forward neural network and summarize our results in Table~\ref{tab:rsquare} and Table~\ref{tab:correlation}. We compare our results to previous state-of-the-art results reported in~\cite{hope2013ploras} and~\cite{hope2018ploras}. We note that, in both previous work, specialized and sophisticated lesion based features were used to obtain the best results. In contrast, our models only make use of either a) raw image features or b) demographic features and raw image features. We observe that compared to previous approaches that use lesion information, our model obtains competitive performance using only image information. However, we note that Hope et al.~2013 and Hope et al.~2018 use different subsets of the dataset. 

\paragraph{Analysis}

\begin{figure}
\centering
\begin{minipage}{.7\textwidth}
  \centering
  \includegraphics[width=0.8\linewidth,keepaspectratio]{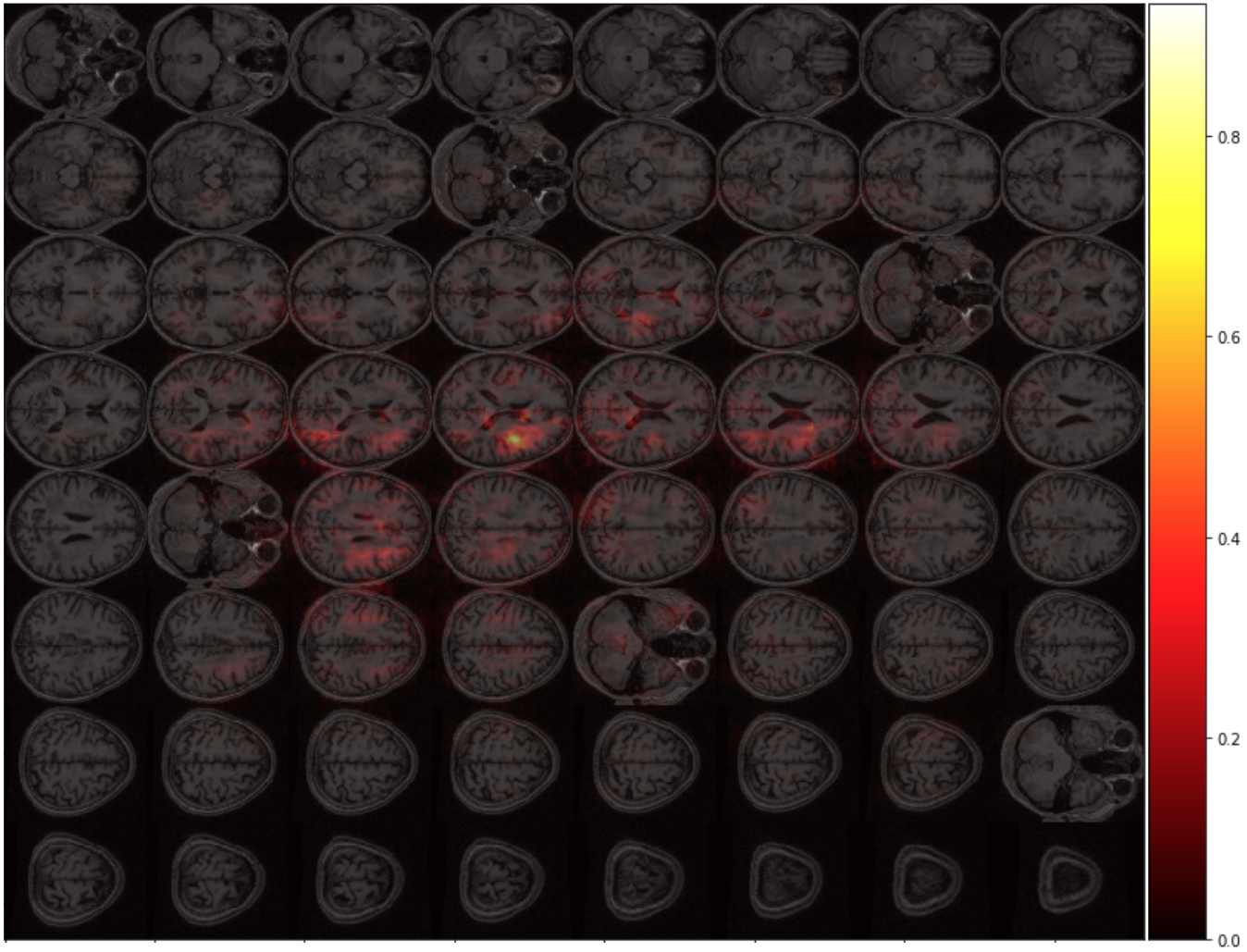}
	\end{minipage}\hfill
	\begin{minipage}{.3\textwidth}
    \centering
    \includegraphics[scale = 0.35]{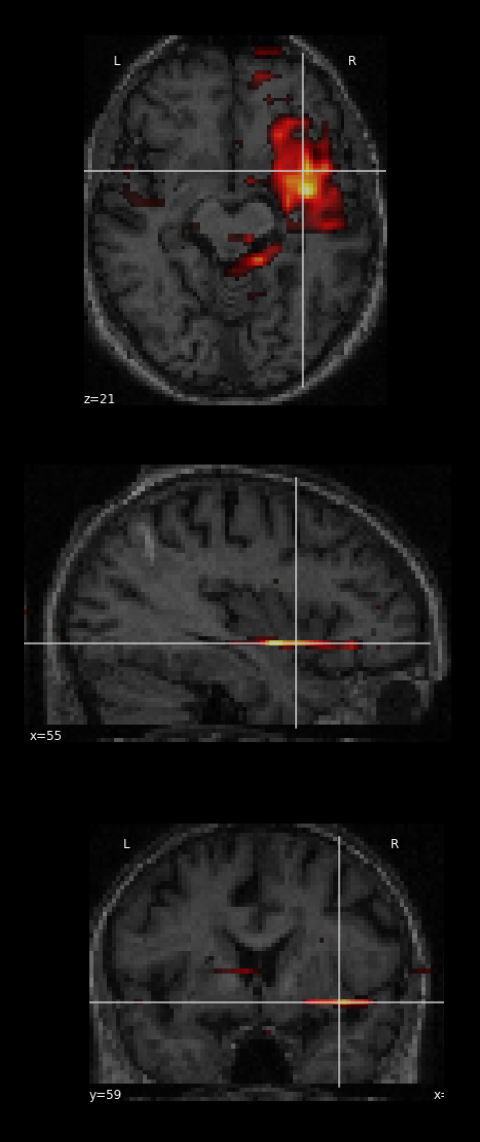}
\end{minipage}\hfill
\caption{Left: Average gradient based saliency maps for the 2D stitched fMRI scans of all patients who scored below the threshold of 60 on the spoken picture description test. Right: Visualization of the same saliency maps in 3-dimensional cross sections (from top: axial, saggital, coronal). Color map is shared and normalized to peak activation.}

\label{fig:saliencies}
\end{figure}

We inspected the flow of gradients within the CNN to identify regions of patient MRI scans that were most salient towards the output. We visualized the regions in the form of a gradient based saliency map averaged across patients belonging to either class (Fig. ~\ref{fig:saliencies} (left)). We further stacked the slices back to compose the original image and visualized the sources of network activation within the original image (Fig. ~\ref{fig:saliencies} (right)). We observe that the right prefrontal region lights up as particularly significant in predicting speech outcome. These results match with a similar study performed to predict reading~\cite{Hoeft361} and language~\cite{hope2017brain} outcomes. We also observe that since the network is trained on 2D horizontal slices, the regions of activation tend to follow a similar format and are dispersed more widely in the axial plane than in the other two. This was an expected constraint given the training methodology. 
We also performed correlation distance analysis~\cite{szekely2007measuring} between the learned representations and the manually extracted white and gray matter and found 
a high distance correlation of \emph{0.7970}. This indicates that the features learned from our proposed CNN based method correlates strongly with the manually extracted gray and white matter features information.

\section{Discussion} 
We have proposed a novel method that makes use of stitched $2$-dimensional plots fed into a convolutional neural network for the prediction of language recovery post stroke. This work provides a preliminary investigation into the utility of breaking $3$-dimensional MRI scans into $2$-dimensional images for the extraction of raw image features that achieve comparable performance to models using more sophisticated information. Our models are able to predict the possibility of recovery competitively, even with very simple CNN based models. Our empirical results indicate that our model learns important representations that are useful for predicting language recovery. Recent work addresses potential challenges in predicting the functional outcome after stroke which may not be entirely captured by an MRI scan~\cite{price2017ten}. Our future work focuses on visualizing the layers and the abstract representations using relevant techniques from computer vision~\cite{zhou2018interpreting,raghu2017svcca}. We would also like to obtain explanations of our predictions using black-box model interpretation techniques~\cite{koh2017understanding,selvaraju2017grad,ribeiro2016should}. 
We are excited by the possibility that the technique might provide a new pathway towards better understanding of language centers and stroke.
%

\subsubsection*{Acknowledgments}
This PLORAS dataset collection was funded by Wellcome (203147/Z/16/Z and 205103/Z/16/Z) and the Stroke Association (TSA PDF 2017/02). We also thank the Alan Turing Institute (EPSRC grant EP/N510129), in particular for hosting the data study group (/TU/B/000012).

\bibliographystyle{unsrt}
\bibliography{refs}
\end{document}